\documentclass[letterpaper]{article} 
\usepackage{aaai2026}  
\usepackage{times}  
\usepackage{helvet}  
\usepackage{courier}  
\usepackage[hyphens]{url}  
\usepackage{graphicx} 
\usepackage{natbib}  
\usepackage{caption} 
\usepackage{algorithm}
\usepackage{algorithmic}

\usepackage{newfloat}
\usepackage{listings}
\DeclareCaptionStyle{ruled}{labelfont=normalfont,labelsep=colon,strut=off} 
\floatstyle{ruled}
\newfloat{listing}{tb}{lst}{}
\floatname{listing}{Listing}
\title{Exploring Depth Generalization in Large Language Models for Solving Recursive Logic Tasks}
\author {
    Zhiyuan He
}
\affiliations{
    University College London\\
    zcabebx@ucl.ac.uk
}

\usepackage{bibentry}

\usepackage{tikz}
\usepackage{multirow}
\usepackage{makecell}
\usepackage{pgfplots}
\usepackage{pgfplotstable}
\usepackage{subcaption}
\usepackage{listings}
\usepackage{amsmath, amssymb}

\usepackage{bibentry}
\pgfplotsset{compat=1.18}
\begin{document}

\maketitle

\begin{abstract}
Large language models have demonstrated remarkable capabilities across many tasks, yet face significant challenges when dealing with recursive reasoning problems, those requiring the resolution of nested hierarchical structures. While prior research has extensively studied length generalization (a model's ability to handle longer sequences than seen during training), we investigate a distinct and underexplored limitation: depth generalization. Here, depth refers to the number of nested levels in a hierarchical problem, such as the layers of parentheses in a mathematical expression or the nesting of logical clauses in a Boolean formula.
Our work reveals that standard transformer architectures struggle with problems involving deeper recursion than encountered during training, even when they perform well on longer but non-nested sequences. This limitation stems from their inability to maintain stack-like behavior, the capacity to track and resolve multiple levels of nested dependencies. Through systematic analysis, we demonstrate how this architectural constraint leads to rapid performance decay as the depth of the recursion increases.
To address this challenge, we develop a novel looped locate-and-replace pipeline that decomposes recursive problems into manageable subcomponents. The approach employs two specialized models: a locator that identifies solvable subexpressions and a replacer that evaluates these components while preserving the overall structure. We evaluated this method in three carefully designed domains: Boolean algebra, recursive arithmetic, and propositional logic, each with a controllable depth of recursion. We show that our method effectively alleviates the performance decay when tested on out-of-distribution recursion depth.
\end{abstract}


\section{Introduction}

Large language models (LLMs), particularly transformer-based architectures \citep{vaswani2017attention}, have achieved remarkable success across diverse domains, from natural language processing to symbolic reasoning \citep{brown2020language, radford2019language}. However, as their adoption grows, understanding their fundamental limitations becomes critical. Recent research has extensively explored the boundaries of transformers, with \textit{length generalization} \citep{lin2025position, xiao2025generalizing, cai2025extrapolation, zhoualgorithms, abbe2024far, JMLR:v25:24-0220, lifunctional, anil2022exploring}, the ability to generalize to longer sequences than those seen during training, being a prominent focus. Tasks such as multi-digit arithmetic, copying, and sorting have served as benchmarks for these studies, revealing both strengths and failures in extrapolating to longer inputs.

Yet, length is only one dimension of generalization. In this work, we investigate a distinct and underexplored axis: \textit{depth generalization}, where the complexity of a problem is measured by its \textit{recursion depth} (e.g., the nesting level of hierarchical structures). While length generalization tests scalability, depth generalization probes a model's capacity for compositional reasoning and handling recursive patterns, a capability central to human cognition and formal systems.
Recursion underpins many fundamental domains, including \textbf{propositional logic} \citep{pospesel1974introduction} (nested quantifiers and clauses), \textbf{Boolean algebra} \citep{boolean} (compound expressions like {\small $(A \land (B \lor \lnot C))$} ), and \textbf{recursive arithmetic} (nested operations like {\small $3 * (2 + (5 / 1))$} ). The depth of recursion reflects the complexity of hierarchical relationships, demanding models to track intermediate states and compose operations systematically. For instance, evaluating an expression with depth $k$ requires resolving $k$ layers of nested dependencies, a challenge distinct from processing a flat sequence of length $k$.

We hypothesize that while non-recursive tasks (e.g., multi-digit addition or sequence reversal) can often be solved via transformers' autoregressive nature or enhanced positional encodings, recursive problems pose a fundamentally harder challenge. Unlike linear sequences, recursive structures require \textit{stack-like behavior}, the ability to push, pop, and backtrack through nested contexts, which transformers lack by design. Attention mechanisms, despite their global receptive field, struggle to implicitly manage dynamic stacks or resolve long-range dependencies across hierarchical layers. This limitation suggests that depth generalization may demand architectural innovations beyond standard positional biases or data scaling, such as explicit memory mechanisms or syntactic scaffolding.

Understanding this gap is essential for applications requiring rigorous symbolic reasoning, such as code generation (recursive function calls) \citep{allamanis2018survey}, automated theorem proving (nested proofs) \citep{irving2016deepmath}, or parsing (syntax trees) \citep{huang2018deep}. By studying depth generalization, we aim to uncover whether transformers can truly \textit{learn recursion} from data or if they require architectural inductive biases to emulate human-like hierarchical reasoning \citep{wei2022symbolic}.

This work aims to bridge the gap in understanding transformers' limitations on depth generalization 
and to develop practical solutions for this underexplored challenge. Below, we outline our contributions:  

\begin{itemize}  
    \item \textit{Diagnose the intuition behind depth generalization failure}:  
    Investigate why transformers struggle with recursion depth (e.g., lack of stack-like mechanisms) compared to length generalization.  
    
    \item \textit{Design an effective pipeline to mitigate recursion depth decay}:  
    Propose a method that enhances transformers' ability to handle nested structures.
    
    \item \textit{Establish benchmarks for depth generalization}:  
    Curate diverse datasets spanning propositional logic, Boolean algebra, and recursive arithmetic to systematically evaluate hierarchical reasoning.  
\end{itemize}  

\section{Related Work}
\label{chap:related-work}

\subsection{Length Generalization in Transformers}
Transformers have shown impressive performance on many tasks, but their ability to generalize to inputs longer than those seen during training remains limited. Early work discovered that while transformers excel at short-sequence tasks like adding two numbers (e.g., ``12+34"), they often fail when given longer inputs (e.g., ``123456+789012") \citep{zhoualgorithms}. Common length generalization benchmarks \citep{lin2025position, xiao2025generalizing, cai2025extrapolation, zhoualgorithms, abbe2024far, JMLR:v25:24-0220, lifunctional, anil2022exploring} include arithmetic operations, sequence copying, and sorting - all of which test how well models can extend their reasoning to longer versions of problems they were trained on. However, these tasks typically focus on \textit{sequential} patterns rather than \textit{hierarchical} ones. For example, reversing a sequence requires processing elements in order, while evaluating a deeply nested mathematical expression requires understanding how operations at different levels interact.

\subsection{Transformers and Recursive Structures}
While length generalization has been well-studied, much less attention has been paid to how transformers handle recursive or nested structures. Recursive problems require models to track hierarchical relationships - like matching parentheses in an expression or resolving nested logical clauses. Research has shown that transformers struggle with even simple recursive patterns like balanced brackets (Dyck languages) \citep{bhattamishra2020dyck}. Interestingly, traditional recurrent neural networks \citep{sherstinsky2020fundamentals} often perform better on these tasks because their architecture naturally supports the "stack-like" operations needed for recursion \citep{deletang2022rnns}. This suggests that the standard transformer architecture may lack crucial inductive biases needed for recursive reasoning.\par
Understanding how transformers process recursive structures has been an active area of research. Mechanistic studies have identified specific attention patterns that resemble stack operations, including the discovery of "deduction heads" that implement tree-climbing operations \citep{brinkmann2024mechanistic} and induction heads that enable copying mechanisms \citep{olsson2022induction}. Other work has traced how information flows through different layers during recursive tasks, revealing parallel processing motifs and depth-bounded recurrent mechanisms \citep{nanda2023progress}. These analyses reveal that while transformers can develop some strategies for handling recursion, they often do so through shortcut algorithms that fail on edge cases rather than true recursive computation \citep{zhang2023recursively}. Our work builds on these insights by systematically measuring depth generalization across multiple recursive tasks and proposing more robust solutions.\par
Recently, Looped Transformer \citep{yang2023looped, fan24looped} was proposed to improve the length generalization by adding the output at each step back to the input tokens at the vector level. It still does not address the depth generalization since it processes at the token level.


\section{Three Recursive Logic Problem Instances}
\label{sec:3instances}

As most research on length generalization, we evaluate the depth generalization on \textit{controlled} and \textit{structured} problems. Specifically, we choose three recursive problems, each with distinct constants, functions, and evaluation goals. To match with the autoregressive nature of Transformer, all problems use postfix notation\footnote{\url{https://simple.wikipedia.org/wiki/Postfix_notation}} (e.g., $e_1 \dots e_k f$) where expressions are evaluated step-by-step from left to right.

The Boolean Algebra evaluation problem operates on truth values where {\small $\mathcal{C} = \{0, 1\}$} represents \textsc{False} and \textsc{True}, with function symbols {\small $\mathcal{F} = \{+, *, -\}$} corresponding to OR, AND, and NOT operations respectively. For example, the postfix expression $1\,0\,+$ evaluates to $1$.
The recursive structure appears when evaluating nested expressions like {\small $0\,1\,*\,1\,+$} which first computes the AND of $0$ and $1$, then ORs the result with $1$.

For the propositional logic truth table computation, we fix two propositions $p$ and $q$ represented as 4-bit truth table encodings: $p = 1100$ and $q = 0101$ where bits correspond to $(T,T), (T,F), (F,T), (F,F)$ valuations. The function set $\mathcal{F} = \{+, *, -, >\}$ includes IMPLY ($>$) alongside Boolean operators. Evaluating $p\,q\,*$ yields $0100$ (the AND truth table), while $p\,q\,>\,-$ computes the negation of material implication resulting in $0010$. This representation preserves the recursive structure of compound formulas while operating on entire truth tables at each step.

The Compositional Arithmetic problem uses 3-digit integer constants $\{000,\dots,999\}$ with three arithmetic operations $\mathcal{F} = \{+, -, *\}$. All operations truncate results to the least significant three digits, making $999\,001\,+$ evaluate to $000$ due to overflow. Nested expressions like $002\,003\,*\,004\,+$ (equivalent to $(2\times3)+4$) demonstrate recursive evaluation.
The problem focuses on recursive depth rather than digit-wise generalization of arithmetic operations.\par
The formal representation of these three problems are included in the supplement materials.


\section{Transformers Fail Depth-Generalization}
Recursive problems require sequential depth-wise computation, which poses a fundamental challenge for transformers due to their fixed-depth architecture. Unlike non-recursive tasks, such as multi-digit addition or multiplication, where positional embeddings, attention mechanisms and auto-regressive generation can effectively handle variable-length inputs, length-generalizing recursive tasks demand an unbounded depth of computation. We argue that transformers are inherently limited in their ability to generalise to recursion depths beyond the number of layers \( L \), as each layer can only process one level of recursion.

To investigate this, we train a GPT-2 model with a binary classification head to evaluate Boolean algebra expressions. The model is trained using Boolean algebra expressions with at most 5 Boolean algebra operations from [AND, OR, NOT], until the loss stops decreasing on the in-length validation set. However, as we will demonstrate, this performance does not generalise to deeper recursion depths. This limitation arises because transformers process information in parallel across layers, while recursion requires sequential depth-wise computation, creating a fundamental mismatch.

\subsection{Comparison of Depth and Length Generalization}
\begin{figure}[htbp]
    \centering
    \includegraphics[width=0.45\textwidth]{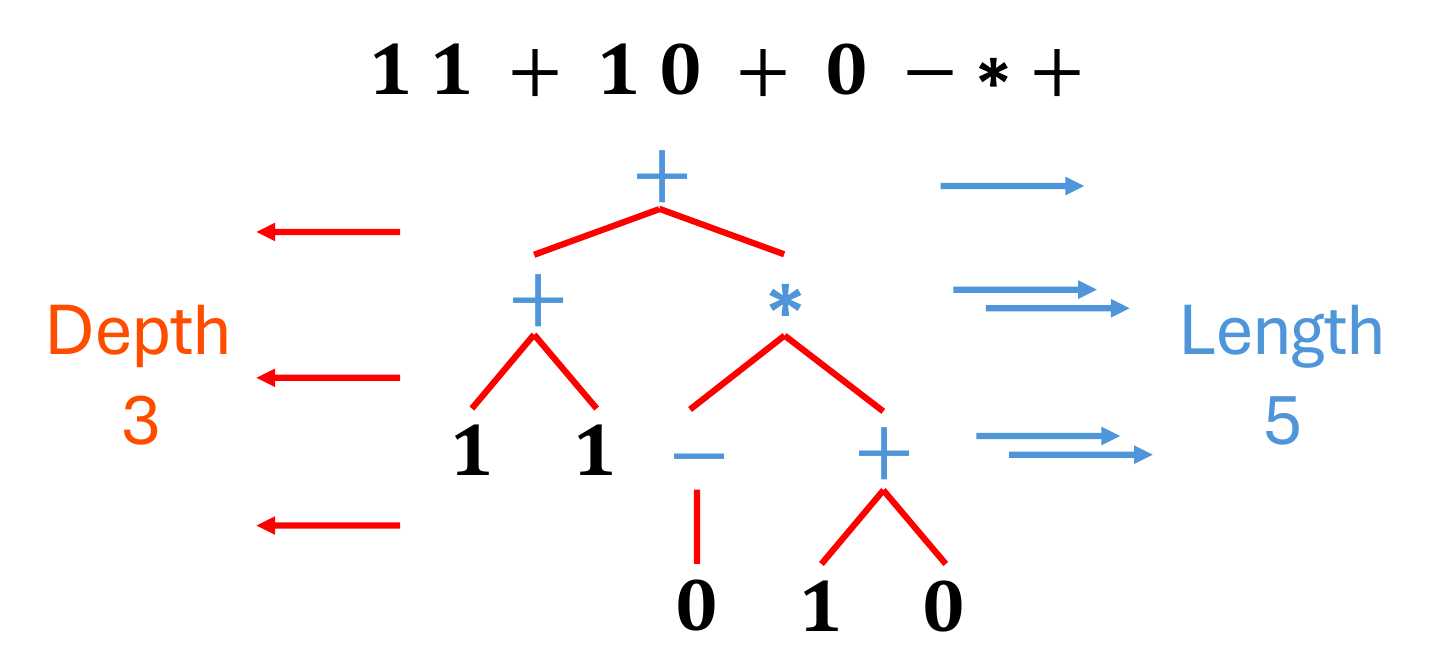}
    \caption{Definition of length and depth.}
    \label{fig:length-n-depth}
\end{figure}

In this work, we distinguish between length and depth when analyzing the generalization behavior of Transformers on recursive logic problems.
We define \textbf{length} as the total number of Boolean algebra operators in an expression, which directly corresponds to the number of computational steps required for full evaluation while \textbf{depth} refers to the depth of the recursion tree of the expression, capturing the hierarchical structure of nested subexpressions. An illustration of this distinction is provided in Figure~\ref{fig:length-n-depth}.

\begin{figure}[]
    \centering
    \includegraphics[width=0.45\textwidth]{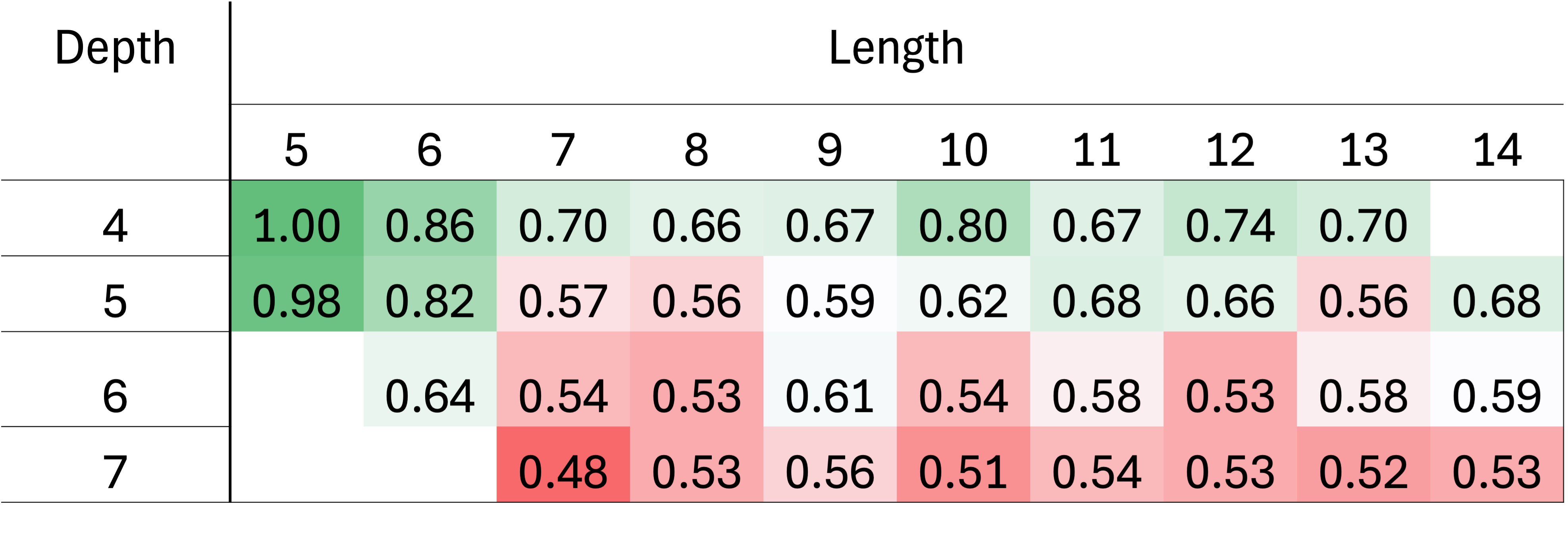}
    \caption{Heatmap of accuracies on Boolean algebra problem of different lengths and depths.}
    \label{fig:heatmap-len-dep}
\end{figure}

We compare the depth and length generalization capabilities of transformers in the task of Boolean algebra evaluation. The out-of-distribution test set is categorized based on the depth and length of the expressions, and the accuracy for each depth-length combination is reported in Figure~\ref{fig:heatmap-len-dep}.
The results indicate that for problems with the same depth, accuracy begins to drop as the length moves just beyond the training distribution. However, as the length further increases, the accuracy does not continue to degrade monotonically but instead fluctuates, showing little correlation with length. 
In contrast, for any given length, the model's performance deteriorates rapidly to approximately 50\% (equivalent to random guessing) as the recursion depth increases. 

\begin{table}[]
    \centering
    \begin{subtable}{0.45\textwidth}
        \centering
        \begin{tabular}{|c|c|}
            \hline
            
            \small \textbf{Depth} & \small \textbf{PCC (Length, Accuracy)} \\
            \hline
            \small 4 & \small -0.3098 \\
            \small 5 & \small -0.1799 \\
            \small 6 & \small -0.1306 \\
            \hline
            \small \textbf{Average} & \small \textbf{-0.2068} \\
            \hline
        \end{tabular}
        \caption{PCC between length and accuracy for depths 4 to 6.}
        \label{tab:depth_corr}
    \end{subtable}
    \hfill
    \begin{subtable}{0.45\textwidth}
        \centering
        \begin{tabular}{|c|c|}
            \hline
            \small \textbf{Length} & \small \textbf{PCC (Depth, Accuracy)} \\
            \hline
            \small 6  & \small -0.9329 \\
            \small 7  & \small -0.9592 \\
            \small 8  & \small -0.8949 \\
            \small 9  & \small -0.8485 \\
            \small 10 & \small -0.9478 \\
            \small 11 & \small -0.9151 \\
            \small 12 & \small -0.9551 \\
            \small 13 & \small -0.8487 \\
            \small 14 & \small -0.9969 \\
            \hline
            \small \textbf{Average} & \small \textbf{-0.9210} \\
            \hline
        \end{tabular}
        \caption{PCC between depth and accuracy for lengths 6 to 14.}
        \label{tab:length_corr}
    \end{subtable}
    \caption{The accuracy has a much
stronger correlation with depth than with length.}
    \label{tab:combined_corr}
\end{table}

We further use the Pearson correlation coefficient (PCC) \citep{cohen2009pearson} to quantify the correlated relationship between accuracy and depth or length, which is defined as
\[
PCC(X, Y) = \frac{\text{Cov}(X, Y)}{\sigma_X \sigma_Y},
\]
where \(\text{Cov}(X, Y)\) is the covariance, \(\sigma_X\) and \(\sigma_Y\) are the standard deviations.

The PCC ranges from \(-1\) to \(1\), where {\small \( 1 \)} and {\small \( -1 \)} indicate a perfect positive and negative relationship, respectively. {\small \( PCC = 0 \)} indicates no relationship.

Table~\ref{tab:combined_corr} show that the average PCC between length and accuracy is -0.2068, while the average PCC between depth and accuracy is -0.9210, showing that accuracy has a much stronger correlation with depth than with length. This suggests that as recursion depth increases, the model's accuracy decreases more significantly and consistently, highlighting a fundamental difficulty in handling deeper recursive structures. Consequently, depth generalization poses a far greater challenge for transformers than length generalization.

\subsection{Layer-Wise Analysis of Transformer Limitations}

\begin{figure}[htbp]
    \centering
        \includegraphics[width=\linewidth]{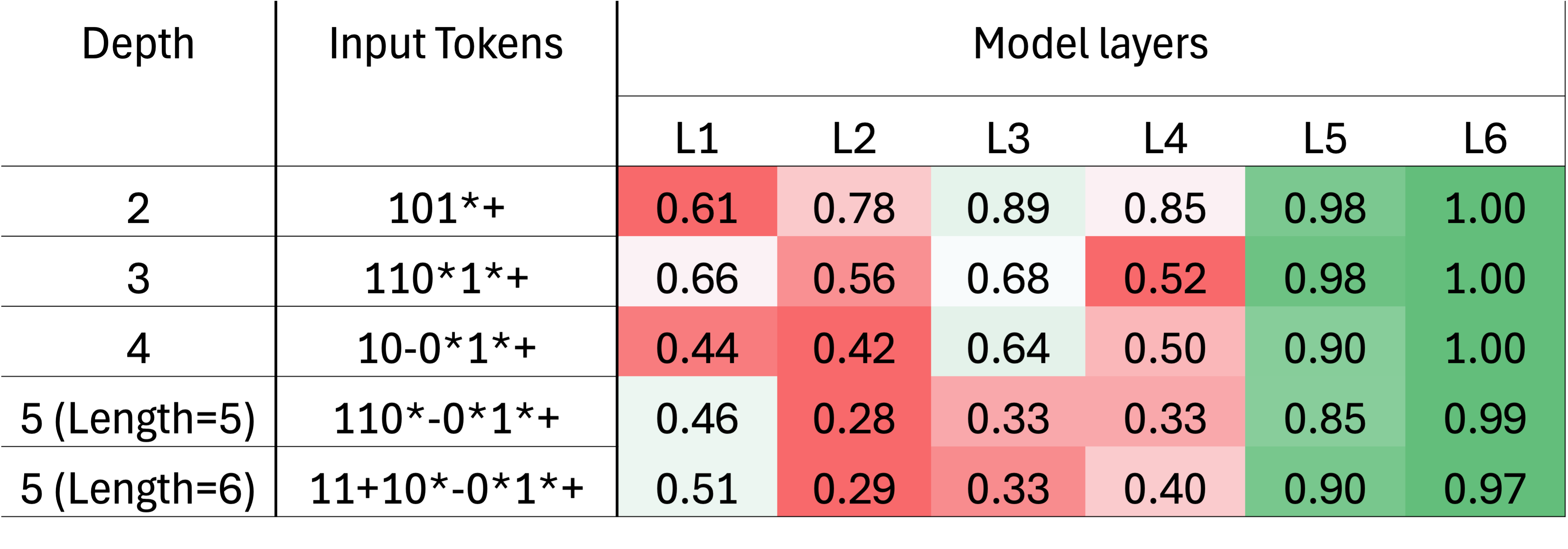}
    \caption{Hidden state visualizations of the last token in Boolean algebra problems with different depth, across different transformer layers. 
    }
    \label{fig:hidden_states_dep}
\end{figure}

Transformers process Boolean algebra expressions recursively, resolving intermediate operations layer by layer. 
The recursion problem must be solved step by step from the bottom to the top in a parse tree, where the intermediate result of each step is only accessible at the next layer. Consequently, solving a problem of depth 3 requires at least 3 layers if we assume that each layer can only resolve one step.

Since transformers have a fixed number of layers, the model struggles to reconstruct intermediate results beyond a certain depth, leading to a rapid performance drop toward 50\% accuracy, equivalent to random guessing.

Also, in shallow cases, local token interactions allow the model to correctly compute sub-expressions. However, as recursion depth increases, the number of intermediate computations grows, requiring hidden states to maintain and propagate information across more layers.

We empirically study how the depth of an operator correlates with the number of layer in which it is evaluated. The first four rows in Figure~\ref{fig:hidden_states_dep} demonstrates how each layer's hidden state prediction changes with problem depth. The hidden state prediction of a layer is calculated by applying the output layer to the hidden state at that layer.

We observe that as depth increases, the correct prediction (in this example, 1) occurs at progressively later layers, suggesting that operations with greater depth are evaluated in later layers.
In contrast, as shown in the last two rows in Figure~\ref{fig:hidden_states_dep}, when depth remains the same but length increases by one, the hidden state prediction remains nearly unchanged. This suggests that the new operation is solved in parallel and does not require additional layers.

\subsection{Limited Impact of Model Size}
\begin{figure}[htbp]
\centering
\begin{tikzpicture}[scale=0.7]
\begin{axis}[
    xlabel={Recursion Depth},
    ylabel={Accuracy (\%)},
    xmin=5, xmax=10,
    ymin=50, ymax=100,
    xtick={5,6,7,8,9,10},
    ytick={50,60,70,80,90,100},
    legend pos=north east,
    grid=major,
    width=12cm,
    height=5cm,
    ymajorgrids=true,
    xmajorgrids=true,
    tick label style={font=\Large},
    label style={font=\Large},
    title style={font=\Large}
]

\addplot[yellow, thick, mark=square*] coordinates {
    (5,100)
    (6,75)
    (7,65)
    (8,60)
    (9,56)
    (10,53)
};

\addplot[orange, thick, mark=triangle*] coordinates {
    (5,100)
    (6,77)
    (7,66)
    (8,61)
    (9,56)
    (10,54)
};

\addplot[red, thick, mark=square*] coordinates {
    (5,100)
    (6,78)
    (7,66)
    (8,62)
    (9,56)
    (10,54)
};

\addplot[purple, thick, mark=triangle*] coordinates {
    (5,100)
    (6,80)
    (7,68)
    (8,62)
    (9,55)
    (10,55)
};

\legend{
    4layer\_4head\_128embd,
    4layer\_4head\_256embd,
    8layer\_8head\_256embd,
    8layer\_8head\_512embd,
}

\end{axis}
\end{tikzpicture}
\caption{Different sized GPT-2 backbone models demonstrate nearly identical performance decay patterns. }
\label{fig:model-size-ablate}
\end{figure}
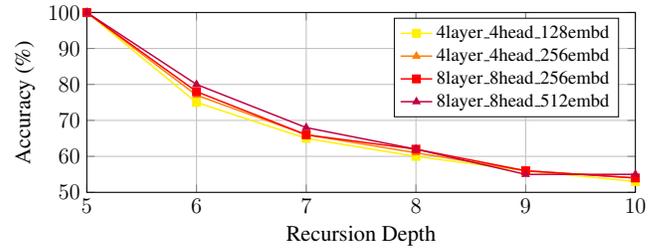

Our experiments reveal that model scale has surprisingly limited impact on depth generalization capabilities. As shown in Figure~\ref{fig:model-size-ablate}, when tested on recursion depths up to 10 (2$\times$ the maximum in-distribution depth), all model variants demonstrate nearly identical performance decay patterns. The accuracy difference between small 4-layer configurations and large 8-layer architectures never exceeds 1\% at any depth level.

The consistent failure modes across scales indicate that transformers primarily rely on surface-level pattern matching rather than developing true recursive reasoning capabilities. Larger models in Figure~\ref{fig:model-size-ablate} show only marginal performance improvements, suggesting they simply memorize more sophisticated patterns without fundamentally changing their approach to hierarchical structures.

The performance plateau with increased model size reveals an inherent architectural limitation. Standard transformers lack the necessary mechanisms for maintaining and resolving recursive dependencies, as evidenced by their similar error profiles on nested Boolean expressions across all model sizes. These results imply that scaling existing architectures may not solve depth generalization challenges, and that explicit architectural modifications such as memory augmentation or syntactic scaffolding may be necessary for tasks requiring genuine recursive reasoning.

\section{Our Approach}
\begin{figure}[htbp]
    \centering
    \includegraphics[width=0.45\textwidth]{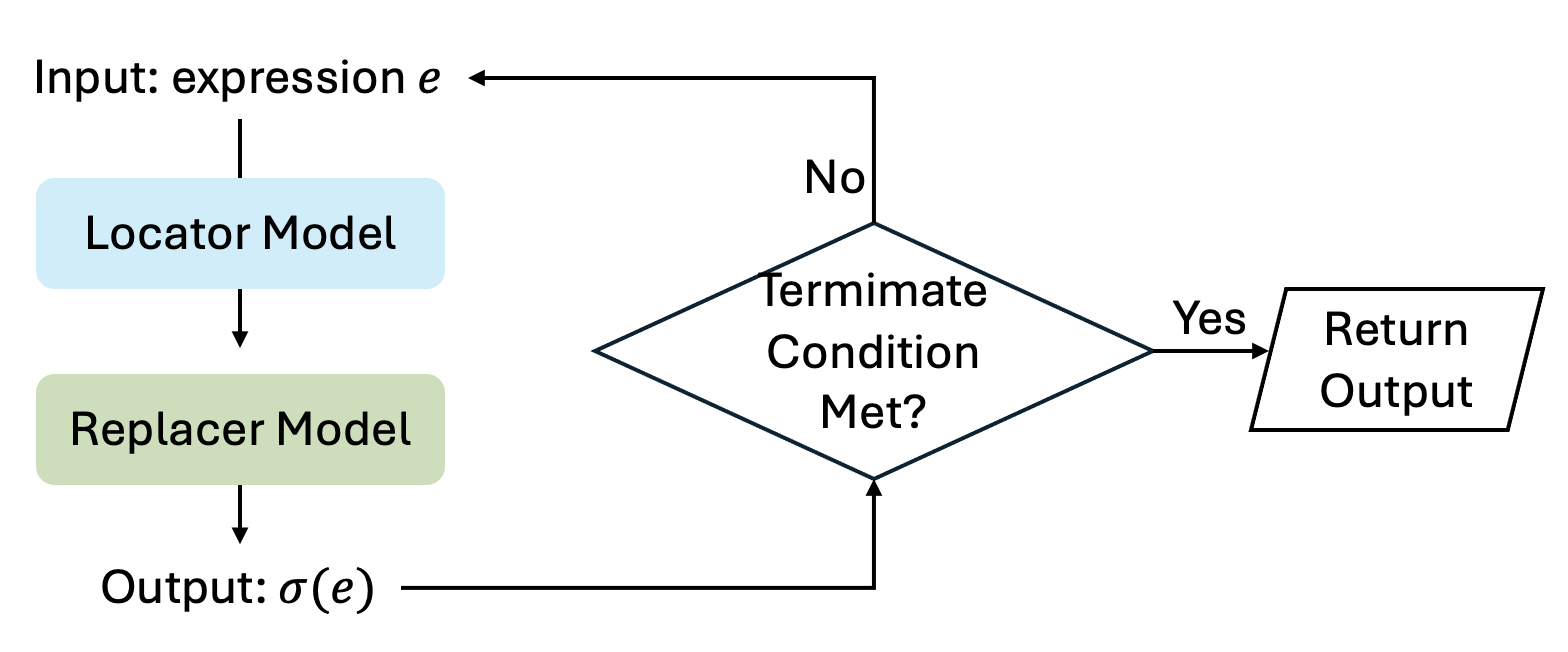}
    \caption{Overall pipeline for the looped locate-and-replace algorithm.}
    \label{fig:full-pipeline}
\end{figure}

\begin{figure*}[htbp]
    \centering
    \includegraphics[width=0.8\textwidth]{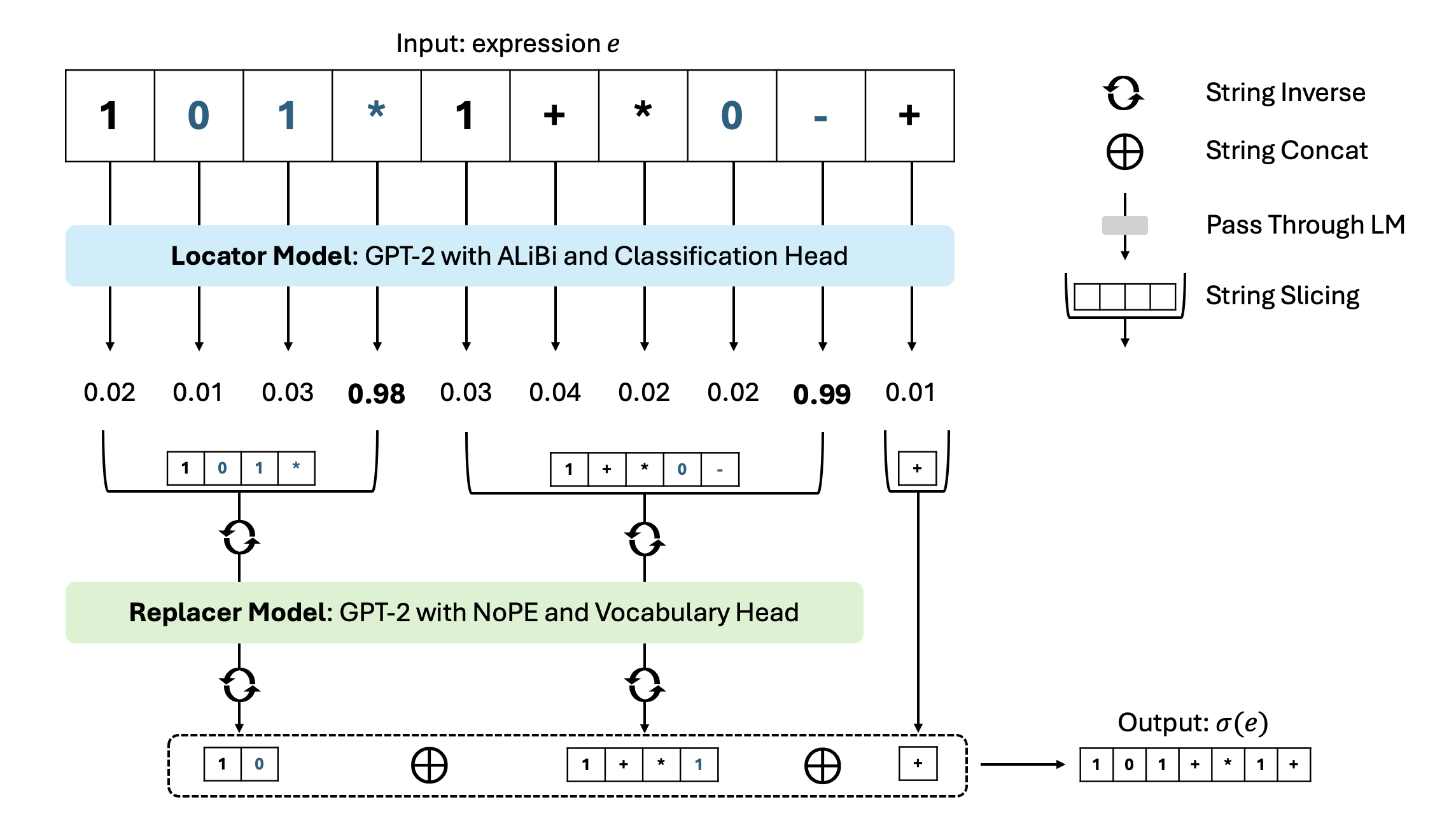}
    \caption{Overall pipeline for the locate-and-replace algorithm.}
    \label{fig:locate-and-replace}
\end{figure*}
To address the challenge of depth generalization, we develop a novel \textit{Looped Locate-and-Replace} (LLR) pipeline, illustrated in Figure~\ref{fig:full-pipeline} and Figure~\ref{fig:locate-and-replace}. The basic ides is to explicitly decomposes recursive problems into manageable subcomponents. For each input expression \(e\), the output of passing the model in one loop is an expression \(\delta(e)\), which is a reduction of \(e\) by computing the direct subexpressions.\par
Inside each loop, the neural network model implements two specialized components: a \textit{locator} that identifies solvable subexpressions and a \textit{replacer} that evaluates these components while preserving the overall structure. In our implementations, both locator and replacer are trained with a same Tranformer-based neural network: GPT-2.\par
While the use of the looped pipeline intuitively has a good match to recursive reasoning problems, the design of locate-replace architecture is based on an analysis that Transformer has fundamental limitation on Match and Replace problems.
\subsection{Limitations of Transformers on Match and Replace Problem}

We analyze the limitations of Transformers on the \textit{Match and Replace Problem} using the \textit{Restricted Access Sequence Processing (RASP)} language \citep{weiss2021thinking}.
RASP is a computational model that abstracts the computation of Transformers into a sequence of operations on input sequences.
It allows us to reason about the expressiveness of Transformers in a formal and precise way.

The \textit{RASP-Generalization Conjecture} \citep{zhoualgorithms} posits that Transformers tend to generalize well to longer input sequences on a task if the task can be solved by a short RASP program that works for all input lengths. In other words, if a task can be expressed concisely in RASP without relying on input-length-specific operations, Transformers are likely to perform well on it across different sequence lengths. Conversely, tasks that require input-length-dependent RASP programs are less likely to generalize well.

The Match and Replace Problem involves identifying a specific subsequence within a given sequence and replacing it according to predefined rules, while copying the non-pattern parts of the sequence unchanged. For example, given the input sequence \texttt{100+*}, the output should be \texttt{10*}, where the pattern \texttt{00+} is replaced with \texttt{0}, and the rest of the sequence (\texttt{1} and \texttt{*}) is copied as is. While this task appears simple, it poses significant challenges for Transformers when implemented in RASP due to the following reasons:

\begin{itemize}
    \item \textbf{Accessing a sequence by a variable is illegal}: 
    To match a pattern in the middle of a sequence, a loop would typically be used, requiring access to the sequence via a variable (e.g., \texttt{seq[i]}). However, this approach is illegal in RASP because Transformers cannot directly access sequence elements using variable indices. Transformers rely on attention mechanisms and positional encodings to process sequences \cite{kazemnejad2023impact}, making variable-based indexing impractical.

    \item \textbf{Non-causal dependencies}: 
    Without using a loop, pattern matching must be performed directly through attention mechanisms. However, matching a pattern that can appear anywhere in the sequence requires either a \texttt{shift\_left} operation (to access future tokens) or attention to tokens on the right side of the current position. Both approaches are forbidden in RASP as they break the causal dependencies required by Transformers during training. A failed RASP program that illustrates this limitation is provided in the supplement materials.
\end{itemize}

According to the RASP-Generalization Conjecture, 
these limitations is fundamental and cannot be overcome without modifying the architecture.
This analysis underscores the importance of developing new architectures that can handle non-causal tasks more effectively, potentially by relaxing the causal constraints or introducing mechanisms for lookahead. Until then, tasks like the Match and Replace Problem will remain challenging for Transformers.

We propose a two-stage locate-and-replace pipeline (Figure~\ref{fig:locate-and-replace})
which consists of two specialized models: (1) a locator model employing ALiBi attention with a binary classification head to detect evaluable patterns, and (2) a replacer model using no positional encoding with a standard language modeling head to generate solutions. This decoupled architecture enables precise pattern localization followed by accurate substitution. 

\subsection{Locator Model}
\label{sec:localisation-method}

To identify the boundaries of an evaluatable base-case sub-expression within a sequence, we employ a locator model that outputs a scalar logit for each token in the sequence.

During the forward pass, the sequence is processed from left to right. For each token, the locator model predicts the probability that an evaluatable base-case pattern ends at that position. 
When the pattern boundary is reached, a distinct probability peak emerges, indicating strong confidence that the evaluable subexpression terminates at this position. 

\subsection{Replacer Model}
After the locator model identifies evaluatable subexpressions, the replacer model computes their evaluations and integrates the results back into the sequence. As shown in Figure~\ref{fig:locate-and-replace}, the locator's markers enable slicing the input string into segments, each terminated by a base-case (except the final segment). The replacer processes each segment independently, generating an output sequence where: (1) base-cases are reduced to their simplified forms, and (2) non-base-case components are copied verbatim. For instance, when processing the segment \texttt{'1+*0-'}, the base-case \texttt{'0-'} evaluates to \texttt{'1'} while \texttt{'1+*'} copies unchanged, producing the intermediate output \texttt{'1+*1'}.

We employ two key optimizations to enhance this process. First, we reverse input sequences (Figure~\ref{fig:locate-and-replace}) to position base-cases before non-evaluatable components (e.g., \texttt{'1+*0-'} $\rightarrow$ \texttt{'-0*+1'}), which significantly improves evaluation accuracy by preventing interference from preceding non-base-case elements. The output is subsequently reversed to restore the original order. Second, following \cite{kazemnejad2023impact}, we use No Positional Encoding (NoPE), which outperforms alternatives like RoPE and ALiBi for copying tasks by leveraging the transformer's pure auto-regressive nature.

The final output is constructed by concatenating all processed segments, completing a full base-case evaluation cycle $\sigma$, as demonstrated in Figure~\ref{fig:locate-and-replace}'s bottom row.

\section{Experiments and Results}
\subsection{Models and Datasets}
\label{sec:models}
The baseline model uses a GPT-2 architecture with reduced dimensions (n\_embd=256, n\_layer=4, n\_head=4) for efficient experimentation.

Three distinct tasks described before were used for evaluation: Boolean algebra evaluation, propositional logic truth table computation, and combinatorial arithmetic. All datasets were generated recursively through automated processes, with the code available in the supplement materials. Each generated sample was classified by its recursion depth to enable depth-specific analysis. The dataset was partitioned into 9,000 training samples, 1,000 validation samples, and 1,000 test samples for each out-of-distribution depth.

\subsection{Training}

In formal logic tasks where precise symbolic manipulation is required, we employ \textbf{character-level tokenization} to ensure robust processing of mathematical expressions. This approach tokenizes each individual character in the input sequence separately, which prevents ambiguous token boundaries that could occur with subword tokenization, and also enables exact positional matching for pattern localization in our locate-and-replace pipeline. 

For each task, we train the model on the problems with recursion at most five. All experiments were conducted with fixed random seeds to ensure reproducibility. The Adam optimizer are used with a learning rate of $5\times10^{-4}$.
Training was performed on NVIDIA A30 GPUs with 24GB memory, using a batch size of 512 which allowed each epoch to complete in approximately 30 seconds. Early stopping was optionally employed when validation loss plateaued. The termination condition for sequence generation was task-dependent: single-character outputs for Boolean Algebra, 4-character outputs for propositional logic, and 3-character outputs for arithmetic. Given our maximum test depth of 12, generation was automatically terminated after 12 recursion iterations to prevent unnecessary computation.

 For the locator model performing binary classification, we use binary cross-entropy loss with logits, defined as
 \[
\mathcal{L}_{\text{loc}} = -\frac{1}{N}\sum_{i=1}^N [y_i\log\sigma(\hat{y}_i) + (1-y_i)\log(1-\sigma(\hat{y}_i))],
\]
where $y_i \in \{0,1\}$ marks evaluable patterns and $\hat{y}_i$ are the logits. The replacer model uses the cross-entropy loss
 \[
 \mathcal{L}_{\text{rep}} = -\sum_{t=1}^T \log p(x_t \mid x_{<t}).
 \]

\subsection{Evaluation and Main Results}

The primary evaluation metric was the accuracy of the full sequence, which required an exact match of the generated output with the ground truth. Performance was analyzed for each depth of recursion to understand how the capabilities of the model scale with the complexity of the problem. 

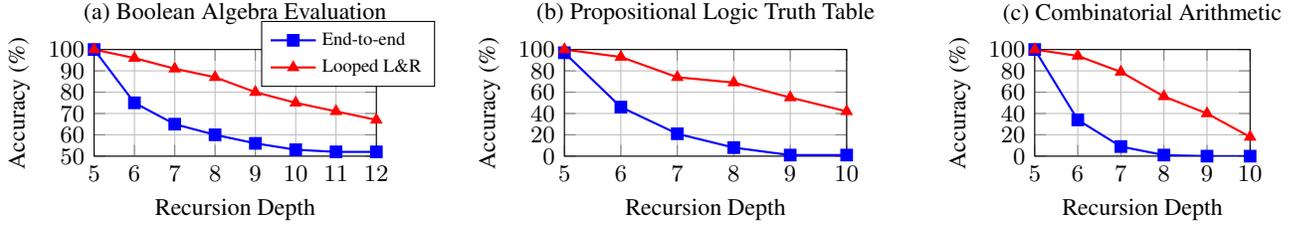
\begin{figure*}[htbp]
\hfill
\begin{subfigure}[b]{0.3\textwidth}
\centering
\begin{tikzpicture}
\begin{axis}[
    xlabel={Recursion Depth},
    ylabel={Accuracy (\%)},
    xmin=5, xmax=12,
    ymin=50, ymax=100,
    xtick={5,6,7,8,9,10,11,12},
    ytick={50,60,70,80,90,100},
    legend pos=north east,
    grid=major,
    width=\textwidth,
    height=3cm,
    ymajorgrids=true,
    xmajorgrids=true,
    title={(a) Boolean Algebra Evaluation},
    title style={font=\small},
    tick label style={font=\small},
    label style={font=\small},
    legend style={
        at={(0.9,0.65)}, 
        anchor=south,     
        font=\scriptsize,
        cells={anchor=west},
        legend columns=1,
    }
]
\addplot[blue, thick, mark=square*] coordinates {
    (5,100) (6,75) (7,65) (8,60) (9,56) (10,53) (11,52) (12,52)
};
\addplot[red, thick, mark=triangle*] coordinates {
    (5,100) (6,96) (7,91) (8,87) (9,80) (10,75) (11,71) (12,67)
};
\legend{
    End-to-end,
    Looped L\&R
}
\end{axis}
\end{tikzpicture}
\end{subfigure}
\hfill
\begin{subfigure}[b]{0.3\textwidth}
\centering
\begin{tikzpicture}
\begin{axis}[
    xlabel={Recursion Depth},
    ylabel={Accuracy (\%)},
    xmin=5, xmax=10,
    ymin=0, ymax=100,
    xtick={5,6,7,8,9,10},
    ytick={0,20,40,60,80,100},
    legend pos=north east,
    grid=major,
    width=\textwidth,
    height=3cm,
    ymajorgrids=true,
    xmajorgrids=true,
    title={(b) Propositional Logic Truth Table},
    title style={font=\small},
    tick label style={font=\small},
    label style={font=\small},
    legend style={
        font=\scriptsize, 
        cells={anchor=west}, 
        legend columns=1, 
        row sep=2pt, 
        column sep=5pt 
    }
]

\addplot[blue, thick, mark=square*] coordinates {
    (5,97) (6,46) (7,21) (8,8) (9,1) (10,1)
};
\addplot[red, thick, mark=triangle*] coordinates {
    (5,100) (6,93) (7,74) (8,69) (9,55) (10,42)
};

\end{axis}
\end{tikzpicture}
\end{subfigure}
\hfill
\begin{subfigure}[b]{0.25\textwidth}
\centering
\begin{tikzpicture}
\begin{axis}[
    xlabel={Recursion Depth},
    ylabel={Accuracy (\%)},
    xmin=5, xmax=10,
    ymin=0, ymax=100,
    xtick={5,6,7,8,9,10},
    ytick={0,20,40,60,80,100},
    legend pos=north east,
    grid=major,
    width=\textwidth,
    height=3cm,
    ymajorgrids=true,
    xmajorgrids=true,
    title={(c) Combinatorial Arithmetic},
    title style={font=\small},
    tick label style={font=\small},
    label style={font=\small},
    legend style={
        at={(1.2,0.5)}, 
        anchor=south,     
        font=\small \footnotesize,
        cells={anchor=west},
        legend columns=1,
    },
]
\addplot[blue, thick, mark=square*] coordinates {
    (5,100) (6,34) (7,9) (8,1) (9,0) (10,0)
};
\addplot[red, thick, mark=triangle*] coordinates {
    (5,100) (6,94) (7,79) (8,56) (9,40) (10,18)
};
\end{axis}
\end{tikzpicture}
\end{subfigure}


\caption{Accuracy decay across recursion depths. The end-to-end transformer performances are marked with blue squares and the looped locate-and-replace (\textit{Looped L\&R}) method performances are marked with red triangles. }
\label{fig:main-result}
\end{figure*}

Figure~\ref{fig:main-result} demonstrates that the proposed looped locate-and-replace (\textit{LLR}) method consistently outperforms the end-to-end vanilla transformer across all three tasks.
The key observation is that the accuracy decay of LLR is significantly milder as recursion depth increases, suggesting better robustness in handling deeper recursive structures.

The end-to-end transformer exhibits a sharp accuracy drop as recursion depth increases, particularly in combinatorial arithmetic where performance quickly approaches near 0\% accuracy at depth 8. In contrast, the method LLR maintains higher accuracy even at greater depths, indicating that stepwise evaluation helps mitigate error accumulation. The gradual decay of LLR may be attributed to the decreasing joint probability of correctness across all depth reduction steps. Interestingly, the decay curve does not follow a strong exponential trend, suggesting that error correlation between steps plays a role.

The Boolean algebra task outputs 0 or 1, presenting a binary classification problem. It means that when accuracy drops to approximately 50\%, the model performs no better than random guessing. The method LLR maintains significantly higher accuracy (66.7\% at depth 12) compared to the end-to-end approach (51.8\%), reinforcing its advantage in structured reasoning tasks.\par
These findings suggest that LLR effectively breaks down complex recursive problems into simpler, more manageable steps, reducing error propagation.

\subsection{Selection of Positional Encoding (PE) Methods}

To address the challenge of base-case solving with depth generalization, we experiment with several PE methods: \textbf{absolute PE}, \textbf{rotary positional embeddings (RoPE)} \citep{su2021roformer}, \textbf{ALiBi (Attention with Linear Biases)} \citep{press2021train}, \textbf{NoPE (no positional encoding)}, and \textbf{inverse-absolute PE}. A detailed description of each method is provided in supplement materials.

\begin{figure}[htbp]
    \centering
    \begin{tikzpicture}
        \begin{axis}[
            xlabel={Problem Depth},
            ylabel={Accuracy},
            ymin=0.5, ymax=1,
            xmin=5, xmax=20,
            xtick={5,8,11,14,17,20},
            ytick={0.5,0.6,0.7,0.8,0.9,1.0},
            legend pos=south east,
            grid=major,
            width=0.45\textwidth,
            height=0.35\textwidth,
            tick label style={font=\small},
            label style={font=\small},
            legend style={
                at={(0.8,0.17)}, 
                anchor=south,     
                font=\scriptsize,
                cells={anchor=west},
                legend columns=1,
            }
        ]

        \addplot coordinates {(5,1.0) (8,0.99) (11,0.99) (14,0.99) (17,0.99) (20,0.99)};
        \addlegendentry{Inverse-Absolute PE}

        \addplot coordinates {(5,1.0) (8,0.97) (11,0.96) (14,0.96) (17,0.96) (20,0.96)};
        \addlegendentry{ALiBi}

        \addplot coordinates {(5,1.0) (8,0.88) (11,0.82) (14,0.82) (17,0.81) (20,0.81)};
        \addlegendentry{NoPE}

        \addplot coordinates {(5,1.0) (8,0.70) (11,0.60) (14,0.57) (17,0.56) (20,0.55)};
        \addlegendentry{Absolute PE}

        \addplot coordinates {(5,1.0) (8,0.72) (11,0.59) (14,0.56) (17,0.56) (20,0.56)};
        \addlegendentry{RoPE}

        \end{axis}
    \end{tikzpicture}
    \caption{Evaluation of positional encoding methods}
    \label{fig:pe-accuracy-plot}
\end{figure}
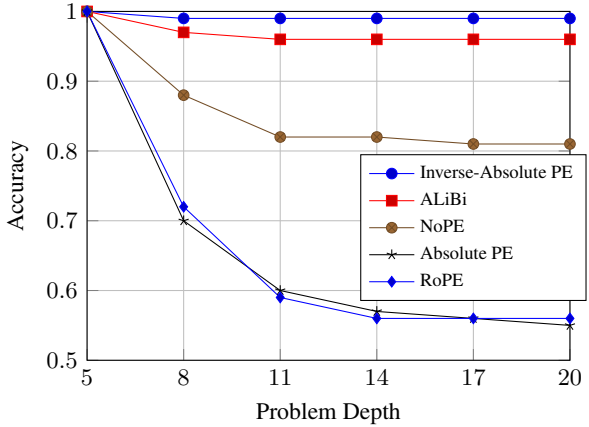

Figure~\ref{fig:pe-accuracy-plot} illustrates the experimental results. When generalized to the length of 20, Inverse-absolute PE achieves the highest accuracy at {\small \textbf{99.9\%}}, followed by ALiBi at {\small \textbf{96\%}}.
Inverse-absolute PE performs best because the base-case pattern always appears at the rightmost position of the input sequence. By assigning a fixed positional embedding to the rightmost token, inverse-absolute PE ensures that the transformer can fit to these fixed patterns, leading to highly accurate base-case matching and replacement. We do not select this method because it does not support auto-regressive generation. In this method, the positional encoding needs to be re-computed when a new token is generated, which makes the forward pass very inefficient. In practice, we select ALiBi as the PE method for the locator model. ALiBi captures the relative positional relationships between tokens, which is beneficial for tasks requiring depth generalization. But it is slightly less robust than inverse-absolute PE because it does not explicitly enforce a fixed positional pattern for the rightmost token. It relies on relative distances, which can introduce variability in how the base-case pattern is processed. 


\subsection{Ablation Study}
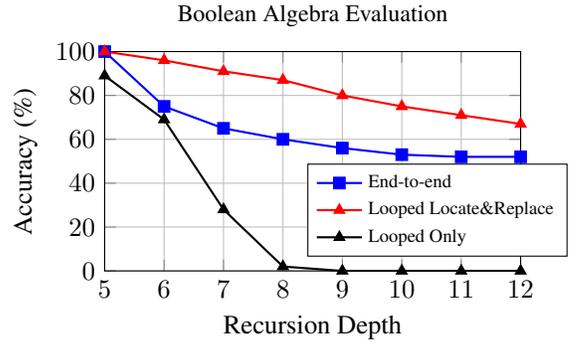
\begin{figure}[ht]
    \centering
\begin{tikzpicture}
\begin{axis}[
    xlabel={Recursion Depth},
    ylabel={Accuracy (\%)},
    xmin=5, xmax=12,
    ymin=0, ymax=100,
    xtick={5,6,7,8,9,10,11,12},
    ytick={0, 20, 40, 60, 80, 100},
    legend pos=north east,
    grid=major,
    width=0.4\textwidth,
    height=4.5cm,
    ymajorgrids=true,
    xmajorgrids=true,
    title={Boolean Algebra Evaluation},
    title style={font=\small},
    legend style={
        at={(0.8,0.065)}, 
        anchor=south,     
        font=\scriptsize,
        cells={anchor=west},
        legend columns=1,
    }
]
\addplot[blue, thick, mark=square*] coordinates {
    (5,100) (6,75) (7,65) (8,60) (9,56) (10,53) (11,52) (12,52)
};
\addplot[red, thick, mark=triangle*] coordinates {
    (5,100) (6,96) (7,91) (8,87) (9,80) (10,75) (11,71) (12,67)
};
\addplot[black, thick, mark=triangle*] coordinates {
    (5,89) (6,69) (7,28) (8,2) (9,0) (10,0) (11,0) (12,0)
};
\legend{
    End-to-end,
    Looped Locate\&Replace,
    Looped Only
}
\end{axis}
\end{tikzpicture}
    \caption{The accuracy of \textit{Looped Only} (black) quickly decays with the increasing depth.}
    \label{fig:ablation-study}
\end{figure}

We implemented a \textit{Looped Only} method without the \textit{Locate} and \textit{Replace}. Figure~\ref{fig:ablation-study} shows that the accuracy of \textit{Looped Only} method quickly decays to zero when the depth increases to eight. Without the \textit{Locate} and \textit{Replace}, the errors from each iteration will accumulate exponentially. These results justify the importance of the \textit{Locate} and \textit{Replace}.

\section{Conclusions}
We investigated depth generalization in transformer-based large language models, a critical yet underexplored axis of generalization that evaluates their ability to handle recursive, hierarchically structured tasks.
Our analysis revealed that standard attention mechanisms struggle to emulate stack-like behavior—essential for resolving deep recursion—leading to rapid performance decay as depth increases.
To address this challenge, we develop a novel looped locate-and-replace pipeline that decomposes recursive problems into manageable subcomponents. We systematically evaluated depth generalization across synthetic datasets, providing empirical insights into failure modes and potential solutions. The findings underscore that the inherent architecture of transformers lacks the inductive biases required for robust recursive reasoning.
By framing depth as a fundamental dimension of generalization, this work lays the groundwork for future research into more structurally aware language models.

\section{Acknowledgments}
The author gratefully acknowledges Professor Anthony Hunter for his supervision and valuable guidance during this project at University College London.

\bibliography{aaai2026}

%

\newtheorem{definition}{Definition}[section]
\newtheorem{theorem}{Theorem}[section]

%
\lstset{%
	basicstyle={\footnotesize\ttfamily},
	numbers=left,numberstyle=\footnotesize,xleftmargin=2em,
	aboveskip=0pt,belowskip=0pt,%
	showstringspaces=false,tabsize=2,breaklines=true}
\floatstyle{ruled}
\newfloat{listing}{tb}{lst}{}
\floatname{listing}{Listing}



\newcommand{\eval}{\operatorname{eval}}
\newcommand{\arity}{\operatorname{arity}}
\newcommand{\Parse}{\operatorname{Parse}}

\makeatletter
\renewenvironment{definition}[1][]{%
  \refstepcounter{definition}%
  \ifx\relax#1\relax
    \def\@tempa{}%
  \else
    \def\@tempa{\textnormal{#1}}%
  \fi
  \trivlist
  \item[\hskip\labelsep \bfseries Definition \thedefinition\@tempa.]%
  \normalfont
}{%
  \endtrivlist
}
\renewenvironment{theorem}[1][]{%
  \refstepcounter{theorem}%
  \ifx\relax#1\relax
    \def\@tempa{}%
  \else
    \def\@tempa{\textnormal{#1}}%
  \fi
  \trivlist
  \item[\hskip\labelsep \bfseries Theorem \thetheorem\@tempa.]%
  \normalfont
}{%
  \endtrivlist
}
\makeatother

\newenvironment{proof}{
PROOF:
\begin{quotation}}{
$\Box$ \end{quotation}}

\newcommand{\nats}{\mbox{\( \mathbb N \)}}
\newcommand{\rat}{\mbox{\(\mathbb Q\)}}
\newcommand{\rats}{\mbox{\(\mathbb Q\)}}
\newcommand{\reals}{\mbox{\(\mathbb R\)}}
\newcommand{\ints}{\mbox{\(\mathbb Z\)}}

%
\pdfinfo{
/TemplateVersion (2026.1)
}

\setcounter{secnumdepth}{0} 

\definecolor{codegreen}{rgb}{0,0.6,0}
\definecolor{codegray}{rgb}{0.5,0.5,0.5}
\definecolor{codepurple}{rgb}{0.58,0,0.82}
\definecolor{backcolour}{rgb}{0.95,0.95,0.92}

\lstdefinestyle{mystyle}{
    backgroundcolor=\color{backcolour},
    commentstyle=\color{codegreen},
    keywordstyle=\color{magenta},
    numberstyle=\tiny\color{codegray},
    stringstyle=\color{codepurple},
    basicstyle=\ttfamily\footnotesize,
    breakatwhitespace=false,
    breaklines=true,
    captionpos=b,
    keepspaces=true,
    numbers=left,
    numbersep=5pt,
    showspaces=false,
    showstringspaces=false,
    showtabs=false,
    tabsize=2
}

\lstset{style=mystyle}

\lstdefinestyle{mystyle}{
    language=Python,
    basicstyle=\ttfamily\small\color{black},  
    keywordstyle=\color{black},               
    commentstyle=\color{gray},                
    stringstyle=\color{black},                
    showstringspaces=false,
    breaklines=true,
    frame=single,
    numbers=left,
    numberstyle=\tiny\color{gray},
    captionpos=b,
    tabsize=4,
    escapeinside={(*@}{@*)},  
}
\pgfplotsset{compat=1.18}

\clearpage
\onecolumn
\section{Appendix}

\subsection{Popular Tasks on Length Generalization}
\begin{table}[ht]
\centering
\begin{tabular}{|lll|}
\hline
\textbf{Task} & \textbf{Input} & \textbf{ Generation} \\
\hline
Multi-digit Addition & \texttt{123 + 456} & \texttt{579} \\
\hline
Multiplication & \texttt{12 * 34} & \texttt{408} \\
\hline
Sequence Copying & \texttt{A B C D} & \texttt{A B C D} \\
\hline
Sequence Reversal & \texttt{A B C D} & \texttt{D C B A} \\
\hline
Sorting & \texttt{1 2 3 4} & \texttt{4 3 2 1} \\
\hline
Dyck & \texttt{((())())} & \texttt{<Valid>} \\
\hline
\end{tabular}
\caption{Length Generalization Tasks and Examples}
\label{tab:length-tasks}
\end{table}

\subsubsection{Formalizations of Recursive Logic Problems}

This section establishes the theoretical foundation for recursive logic problems, beginning with their mathematical formalization. We first present the core definition that characterizes these problems (Section 3.2.1), followed by an analysis of their fundamental properties (Section 3.2.2). Building on this formalism, Section 3.3 then introduces three concrete problem instances - Boolean algebra evaluation, propositional logic truth tables, and combinatorial arithmetic - which serve as our experimental testbed while demonstrating the broad applicability of the framework.

\label{sec:formalization}

\subsubsection{Core Definition}

\begin{definition}
A \textit{recursive problem} is a 6-tuple $\mathcal{R} = \langle \mathcal{C}, \mathcal{F}, \mathcal{B}, \sigma, \delta, \ell \rangle$ where:

\begin{enumerate}
    \item $\mathcal{C}$ is a countable set of constants

    \item $\mathcal{F}$ is a finite set of function symbols where each $f \in \mathcal{F}$ has:
    \[
    \arity(f) = k \quad \text{and} \quad f \colon \mathcal{C}^k \to \mathcal{C}
    \]

     We further define an expression $e$ recursively by the following grammar:
    \[
    \begin{aligned}
    e &::= c \mid  e_1 \dots e_k f 
    \end{aligned}
    \]
    where $c \in \mathcal{C},  f \in \mathcal{F}, k = \arity(f)$

    \item $\mathcal{B}$ is the set of base cases, consisting of all expressions of the form:
    \[
    c_1 \dots c_k f \quad \text{where} \quad c_i \in \mathcal{C} \text{ and } k = \arity(f)
    \]

    \item $\sigma$ is the base-case evaluation rule which solves all the base-cases in an expression and reduce the depth by one. For any expression $e$:

    \[
    \sigma(e) = \begin{cases}
    c & \text{if } e = c \in \mathcal{C} \\
    f(c_1, \dots, c_k) & \text{if } e = c_1 \dots c_k f \in \mathcal{B} \\
    \sigma(e_1) \dots \sigma(e_k)f & \text{if } e = e_1 \dots e_k f \notin \mathcal{B}
    \end{cases}
    \]

    We further define $\sigma^*$ as the recursive evaluation rule which recursively solves the base-cases in an expression until depth is reduced to 0:

    \[
    \sigma^*(e) = \begin{cases}
    c & \text{if } e = c \in \mathcal{C} \\
    
    f(c_1, \dots, c_k) & \text{if } e = c_1 \dots c_k f \in \mathcal{B} \\
    
    f(\sigma(e_1), \dots, \sigma(e_k)) & \text{if } e = e_1 \dots e_k f \notin \mathcal{B}
    \end{cases}
    \]
    where $c_* \in \mathcal{C}$

    \item $\delta$ computes recursion depth:
    \[
    \delta(e) = \begin{cases}
    0 & \text{if } e \in \mathcal{C} \\
    1 + \max\limits_{1 \leq i \leq k} \delta(e_i) & \text{if } e = e_1 \dots e_k f \\ \quad\quad\quad \text{ and } k = \arity(f)
    \end{cases}
    \]
    
    \item $\ell$ counts expression length:
    \[
    \ell(e) = \begin{cases}
    0 & \text{if } e \in \mathcal{C} \\
    1 + \sum\limits_{i=1}^k \ell(e_i) & \text{if } e = e_1 \dots e_k f \\ \quad\quad\quad \text{ and } k = \arity(f)
    \end{cases}
    \]
\end{enumerate}

\end{definition}

\subsubsection{Core Properties}

\begin{theorem}[    \textbf{Evaluation Termination}]
For any expression $e \in \mathcal{E}$, the evaluation function $\sigma^*(e)$ terminates in finite steps, as:
\begin{enumerate}
    \item Base case evaluation terminates immediately when $e \in \mathcal{C} \cup \mathcal{B}$
    \item Base-case evaluations strictly reduce the depth $\delta(\sigma(e)) < \delta(e)$ 
    
\end{enumerate}
Formally, $\forall e \in \mathcal{E},\ \exists n \in \mathbb{N}$ such that $\sigma^*(e)$ completes in $n$ steps.
\end{theorem}

\begin{theorem}[    \textbf{Unambiguous Parsing}]
Assume that the constants are atomic and unambiguously distinguishable, for any postfix expression $e$, there exists exactly one parse tree satisfying:
\[
\Parse(e) = \begin{cases}
c & \\\quad \quad \text{if } e = c \in \mathcal{C} \\
f(\Parse(e_1), \dots, \Parse(e_k)) & \\ \quad\quad \text{if } e = e_1 \dots e_k f  \text{ and } k = \arity(f)
\end{cases}
\]
\end{theorem}

\begin{theorem}[    \textbf{Depth-Length Relationship}]
For any expression $e$:
\[
\delta(e) \leq \ell(e) \leq |\mathcal{F}|^{\delta(e)}
\]

\end{theorem}

\clearpage

\subsection{Three Recursive Problems}

\begin{table}[H]
\centering
\renewcommand{\arraystretch}{1.5} 
\begin{tabular}{|c|c|c|c|}
\hline

\textbf{Components} & \textbf{Boolean} & \textbf{Propositional} & \textbf{Arithmetic} \\
\hline

\makecell{constants\\$\mathcal{C}$} & \makecell{\\$1$\\\\$0$} & \makecell{\\$0000$\\\\$0001$\\\\$0010$\\\\...\\\\$1111$\\\\} & \makecell{\\$000$\\\\$001$\\\\$002$\\\\...\\\\$999$\\\\} \\
\hline

\makecell{functions\\$\mathcal{F}$} & \makecell{\\$+^2$\\\\$*^2$\\\\$-^1$\\\\}
& \makecell{\\$+^2$\\\\$*^2$\\\\$-^1$\\\\$>^2$\\\\} & \makecell{\\$+^2$\\\\$-^2$\\\\$*^2$\\\\} \\
\hline

\makecell{base-cases\\$\mathcal{B}$} & \makecell{\\$11+$\\\\$10*$\\\\$0-$\\\\...\\\\}
& \makecell{\\$11000101>$\\\\$10100011+$\\\\$10001110*$\\\\...\\\\} 
& \makecell{\\$453857+$\\\\$053007*$\\\\$403857-$\\\\...\\\\}  \\
\hline

\makecell{recursion depth\\$\delta$} & 
\makecell{\\$\delta(11+10**)=2$\\\\$\delta(110**)=2$\\\\$\delta(10*)=1$\\\\...\\\\} & 
\makecell{\\$\delta(110011001000+*)=2$\\\\$\delta(11001011>)=1$\\\\...\\\\} & 
\makecell{\\$\delta(222111+003*)=2$\\\\$\delta(333003*)=1$\\\\...\\\\} \\
\hline

\makecell{expr. length\\$\ell$} & 
\makecell{\\$\ell(11+10**)=3$\\\\$\ell(110**)=2$\\\\$\ell(10*)=1$\\\\...\\\\} & 
\makecell{\\$\ell(110011001000+*)=2$\\\\$\ell(11001011>)=1$\\\\...\\\\} & 
\makecell{\\$\ell(222111+003*)=2$\\\\$\ell(333003*)=1$\\\\...\\\\} \\
\hline

\makecell{\\expression\\example\\\\} & $10+0-*1*$ & \texttt{01+11-01>} & \texttt{034051+003*} \\
\hline

\end{tabular}
\caption{Recursive Problem Instances}
\label{tab:instance_repr}
\end{table}

\begin{table*}[htbp]
\centering
\renewcommand{\arraystretch}{1.5} 
\begin{tabular}{|c|c|c|}
\hline

\multirow{2}{*}{\textbf{Problem}} & \multicolumn{2}{c|}{\textbf{Evaluation Methods}} \\ \cline{2-3}
                                  & \textbf{Base-Case Evaluation $\sigma$} & \textbf{Recursive Evaluation $\sigma^*$} \\ \hline
Boolean & \makecell{\\$\sigma(1\underline{11+}0**)=1\underline{1}0**$\\\\$\sigma(1\underline{1-}1+0**)=1\underline{0}1+0**$\\\\$\sigma(1\underline{1-}+\underline{10*}*)=1\underline{0}+\underline{0}*$\\...\\} & 
\makecell{\\$\sigma^*111+0**)=0$\\\\$\sigma^*(11-1+0**)=0$\\\\$\sigma(11-+10**)=0$\\...\\}   \\ 
\hline

Prop. & 
\makecell{\\$\sigma(1100\underline{11001000+}*)=1100\underline{1100}*$\\\\ $\sigma(1100\underline{11001000>}*)=1100\underline{1111}*$\\\\
$\sigma(1100\underline{11001000*}>)=1100\underline{1000}>$ \\...\\} &  
\makecell{\\$\sigma(110011001000+*)=1100$\\\\ $\sigma(1100{11001000>}*)=1100$\\\\
$\sigma(1100{11001000*}>)=1111$ \\...\\} 
\\ \hline

Arithm. & \makecell{\\$\sigma(222111+003*)=333003*$\\\\
$\sigma(222111+666333-*)=333333*$\\...\\} & 
\makecell{\\$\sigma^*(222111+003*)=999$\\\\ 
$\sigma(222111+666333-*)=889*$\\...\\}   \\ 

\hline
\end{tabular}
\caption{Evaluation Examples of Recursive Problem Instances}
\label{tab:instance_eval}
\end{table*}

\clearpage
\subsection{Abstraction of Layer Processing}
\begin{figure*}[htbp]
    \centering
    \includegraphics[width=0.8\textwidth]{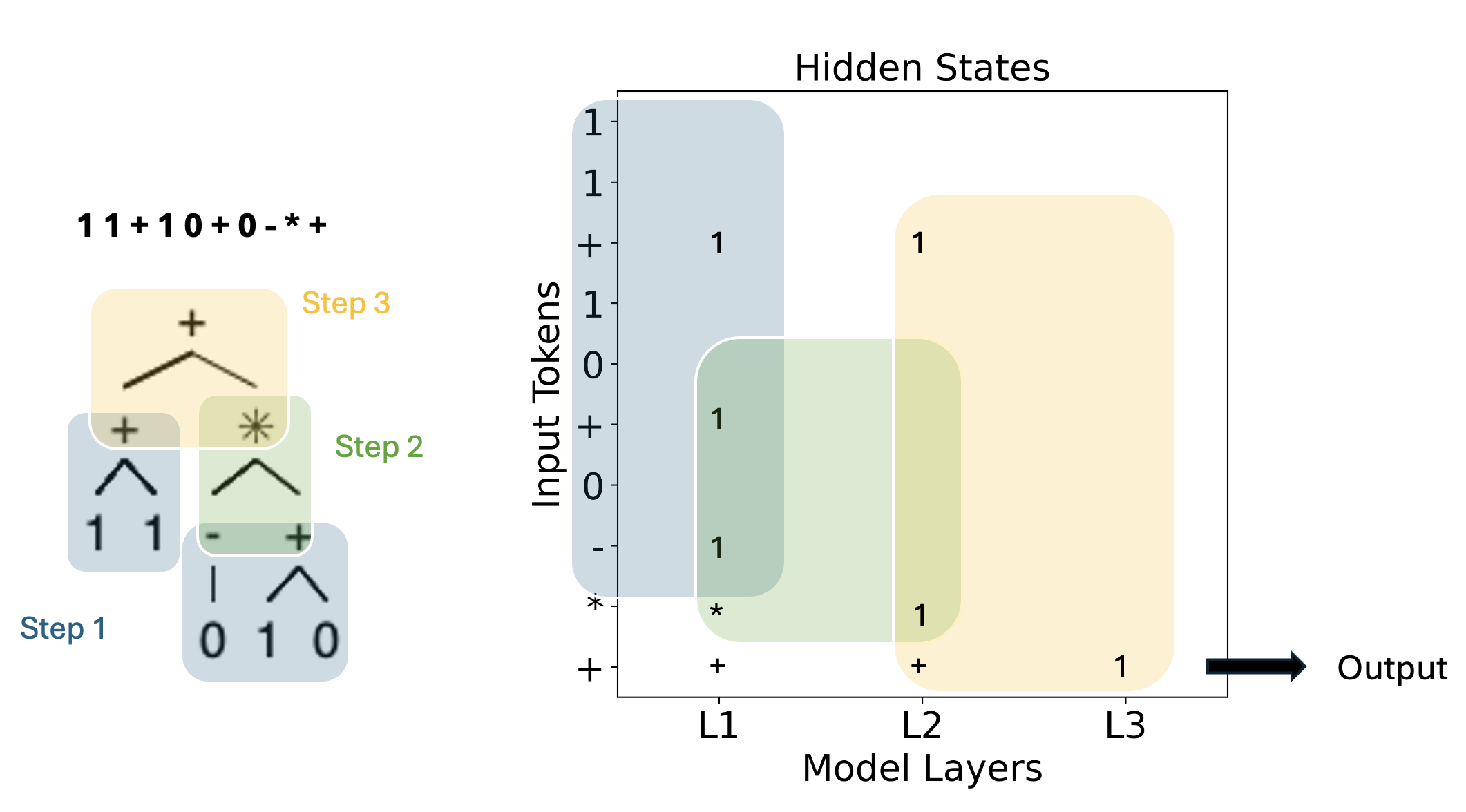}
    \caption{Abstraction of how the computation at each recursive solved layer by layer.}
    \label{fig:layer-demo}
\end{figure*}

\subsection{Illegal RASP-L Example}
\begin{lstlisting}[
    style=mystyle, 
    caption={RASP-L implementation of matching and replacing the pattern '00+'. It is illegal because the attention calcualtion uses the future tokens which breaks the non-causal dependency.}, 
    label=lst:match_replace,
    escapeinside={(*@}{@*)}  % Enable LaTeX escape inside the code block
]
def match_and_replace(seq):
    # Create a sequence of indices for the input sequence
    idx = indices(seq)

    # Create a boolean mask for the first '0' in the pattern '00+'
    is_first_zero = seq_map(seq, full(seq, '0'), equals)

    # Shift the sequence right by 1 to check the next element
    # (*@\textcolor{red}{This is ILLEGAL because it depends on future elements (non-causal).}@*)
    shifted_seq = shift_right(seq, 1, default='')  
    is_second_zero = seq_map(shifted_seq, full(seq, '0'), equals)

    # Shift the sequence right by 2 to check the element after the two '0's
    # (*@\textcolor{red}{This is ILLEGAL because it depends on future elements (non-causal).}@*)
    shifted_seq_plus = shift_right(seq, 2, default='') 
    is_plus = seq_map(shifted_seq_plus, full(seq, '+'), equals)

    # Combine the masks to find the start of the pattern '00+'
    pattern_start = seq_map(is_first_zero, is_second_zero, lambda x, y: x and y)
    pattern_start = seq_map(pattern_start, is_plus, lambda x, y: x and y)

    # Create a mask for the positions to replace with '0'
    replace_mask = seq_map(pattern_start, shift_right(pattern_start, 1, default=False), lambda x, y: x or y)
    replace_mask = seq_map(replace_mask, shift_right(pattern_start, 2, default=False), lambda x, y: x or y)

    # Replace the pattern '00+' with '0'
    # Use the `where` function to conditionally replace elements
    result = where(replace_mask, full(seq, '0'), seq)

    return result
\end{lstlisting}

\clearpage
\subsection{More discussion on Positional Encoding Method Selection}
Absolute PE assigns a unique embedding to each position in the sequence, but it struggles with depth generalization as it does not explicitly encode relative distances between tokens. RoPE incorporates relative positional information by applying a rotation matrix to the token embeddings based on their positions, but it may not be optimal for our task due to its focus on relative distances rather than fixed positional patterns. ALiBi introduces a linear bias to the attention scores based on the relative distances between tokens, explicitly encoding relative positions to improve depth generalization. NoPE removes positional encoding entirely, relying solely on the autoregressive nature of the model, which makes it heavily influenced by the leading tokens and lacks explicit positional information. Inverse-absolute PE assigns positional embeddings in reverse order, ensuring that the rightmost token always receives a fixed positional embedding, making it particularly suited for our task where the base-case pattern always appears at the rightmost position.
In our experiments, Inverse-absolute PE performs best because the base-case pattern always appears at the rightmost position of the input sequence. By assigning a fixed positional embedding to the rightmost token, inverse-absolute PE ensures that the transformer can fit to these fixed patterns, leading to highly accurate base-case matching and replacement. We do not select this method because it does not support auto-regressive generation. In this method, one needs to re-compute the positional encoding when a new token is generated, which makes the forward pass very inefficient. In practice, we select ALiBi to be the positional encoding method for the locator model. ALiBi captures the relative positional relationships between tokens, which is beneficial for tasks requiring length generalization, but it is slightly less robust than inverse-absolute PE because it does not explicitly enforce a fixed positional pattern for the rightmost token. Instead, it relies on relative distances, which can introduce variability in how the base-case pattern is processed. 

NoPE relies entirely on the autoregressive nature of the model, making it more sensitive to the leading tokens in the sequence. Without explicit positional information, the model struggles to consistently identify and replace base-case patterns, especially when the sequence length varies.

Both RoPE and absolute PE perform poorly in our task, achieving only around \textbf{56\%} accuracy. The failure of absolute PE can be attributed to its inability to generalize well to sequences longer than those seen during training and its lack of relative positional relationships, which are crucial for identifying base-case patterns. RoPE, while incorporating relative positional information, may not be well-suited for tasks where the base-case pattern always appears at a fixed position (the rightmost position). The rotation-based encoding introduces variability that can hinder the model's ability to overfit to specific positional patterns. These findings highlight the importance of selecting the right positional encoding method for tasks requiring precise pattern matching and replacement.

\end{document}